\documentclass[conference]{IEEEtran}
\IEEEoverridecommandlockouts
\usepackage{cite}
\usepackage{amsmath,amssymb,amsfonts}
\usepackage{graphicx}
\usepackage{textcomp}
\usepackage{multirow}
\usepackage{booktabs}
\usepackage{epstopdf}
\usepackage[dvipsnames]{xcolor}
\usepackage{url}
\usepackage[hidelinks]{hyperref}
\usepackage{threeparttable}
\usepackage[utf8]{inputenc}
\usepackage[english]{babel}
\usepackage[belowskip=-1pt,aboveskip=0pt]{caption}
\def\BibTeX{{\rm B\kern-.05em{\sc i\kern-.025em b}\kern-.08em
T\kern-.1667em\lower.7ex\hbox{E}\kern-.125emX}}
\begin{document}
\title{\LARGE{\textbf{Multi-Scale Fruit Capsules: Dilated Convolutions and Dynamic Routing for In-the-Wild Explainable Fruit Recognition}}}
\author{\IEEEauthorblockN{Subhankar Chattoraj$^\ddagger$, Sawon Pratiher$^\S$, Samiran Das$^\P$ and Hubert Konik$^\dagger$}
\IEEEauthorblockA{$^\ddagger$\textit{Dept. of Information Systems, University of Maryland Baltimore County, Baltimore, MD 21250, USA}\\
$^\S$\textit{Dept. of Electrical Engineering, Indian Institute of Technology Kharagpur, West Bengal 721302, India}\\
$^\P$\textit{Dept of Data Science and Engineering, Indian Institute of Science Education and Research Bhopal 462066, India}\\
$^\dagger$\textit{UJM-Saint-Etienne, CNRS, Institut d'Optique Graduate School, Laboratoire Hubert Curien, Saint-Etienne, France}\\
\thanks{\textit{{*}Corresponding authors: Subhankar Chattoraj and Sawon Pratiher (e-mail: chattorajsubhankar@gmail.com; sawon1234@gmail.com)}}}}
\maketitle
\thispagestyle{empty}
\pagestyle{empty}
\begin{abstract}
The same fruit appears in a bunch, unpicked, peeled, bagged in plastic, or sliced on a dish, so automated fruit classification in the wild (AFCW) must absorb wide intra-class and narrow inter-class variability in shape, size, colour and texture. Convolutional networks route information through pooling, which discards the pose and location of the region of interest and therefore generalises poorly across these presentations. We propose \textbf{FruitCapsNet}, a capsule network whose Fruit Capsules replace the standard convolutional front end with dilated convolutions: the receptive field grows exponentially at constant parameter cost, so each capsule encodes multi-scale context before dynamic routing resolves part whole spatial agreement. Hyper-parameters, including the dilation factor, are selected by Bayesian optimisation rather than grid search. On three public datasets (SMP, FruitsGB, Fruits-360) and a new 19-class, 10{,}639-image in-the-wild dataset (PD-19), FruitCapsNet exceeds ten fine-tuned transfer-learning backbones at one-third the depth, with the largest margin (+2.7\% over the nearest competitor) on the hardest set. Grad-CAM saliency propagated from the DigitCaps layer shows that the improvement comes from attributing decisions to whole-fruit regions rather than to object edges, giving post-hoc evidence that the gain is not a dataset artefact.
\end{abstract}
\section{Introduction}
Fruit and vegetable processing is a high-volume, low-margin industry in which grading, sorting and cataloguing are still largely manual: labour-intensive, expert-dependent and increasingly hard to staff \cite{Intro_1},\cite{Intro_5}. Automating categorisation, ripeness sorting and defect detection with commodity cameras is therefore an active requirement for IoT-enabled farms, packing lines and retail checkout \cite{Intro_6},\cite{Intro_8},\cite{Intro_9}. The vision problem is harder than the benchmark accuracies suggest. Fruits vary in size, shape, colour and texture with ripening stage, season and provenance, and they are photographed peeled, sliced, washed, bagged or still on the branch \cite{Intro_10}. The result is wide intra-class variation coupled with narrow inter-class separation, which low-level descriptors cannot resolve \cite{Intro_11}. Alternative sensing modalities sidestep the appearance problem but not the cost problem: chemical sensors are invasive and consume the sample, electronic-nose and near-infrared rigs are more expensive than high-resolution cameras, hyperspectral acquisition is slow, and thermal imaging fails when the fruit-background temperature gradient is small \cite{Intro_13},\cite{Intro_16},\cite{bioucas2012hyperspectral},\cite{Intro_15}. Colour imaging remains the only economically deployable option, so the burden falls on the model.

\begin{figure}[!t]
\centering
\includegraphics[width=\columnwidth, height=2.5cm]{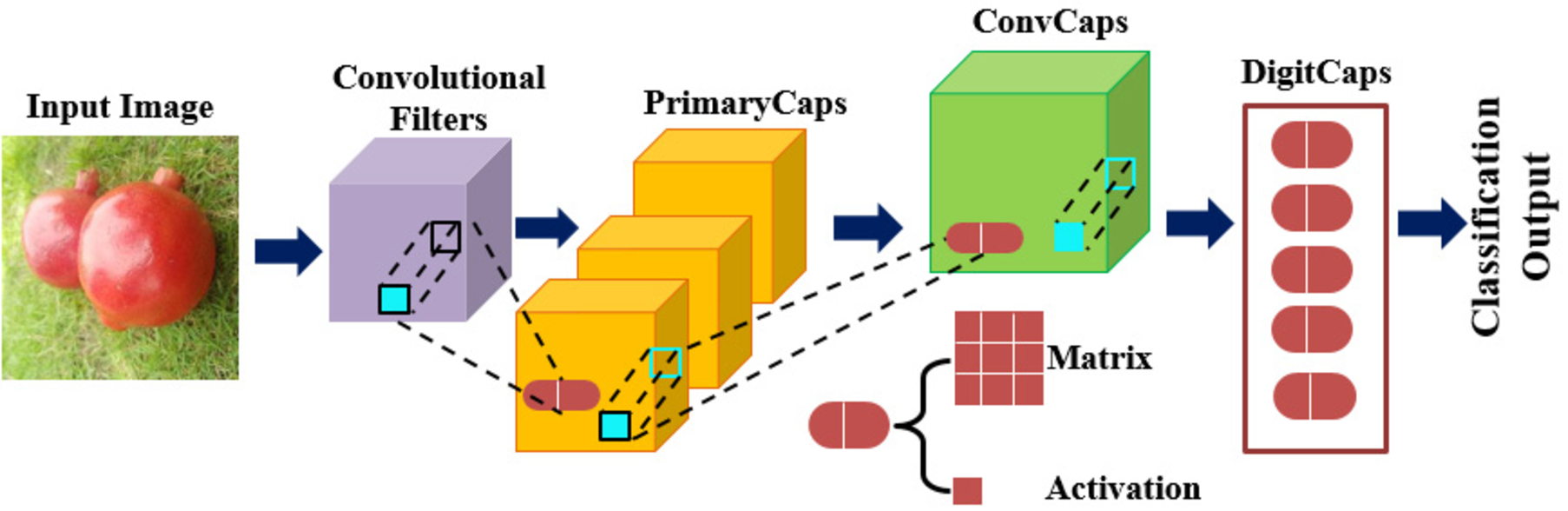}
\caption{Block diagram of the FruitCapsNet-based AFCW.}
\label{fig_pipeline}
\vspace{-2mm}
\end{figure}

Deep learning has demonstrated remarkable success across diverse real-world domains, ranging from medical imaging \cite{ref_26, Ref_27, kheiralipour2017introducing, das2019covariance, das2019sparsity} and renewable-energy systems \cite{Chattoraj2021ImprovingSO, Banerjee2020DeepRL} to industrial automation \cite{Das2024HydrophobicityBasedGO} and intelligent manufacturing. In computer vision, these advances have enabled increasingly robust solutions for image classification, object detection, and visual recognition under complex real-world conditions. Convolutional neural networks (CNNs) dominate industrial imaging \cite{Deep_Learning_1},\cite{Deep_Learning_2} \cite{Das2024HydrophobicityBasedGO} but have a structural weakness for this task. Their routing mechanism, average or max pooling, records that a feature is present in some region of interest (RoI) while discarding where it is and how it is posed, giving translation invariance and only augmentation-induced rotational invariance \cite{Deep_Learning_3},\cite{Deep_Learning_4},\cite{sabour2017dynamic}. A CNN therefore treats an object as a bag of patches and cannot represent the position of one part relative to another, exactly the information that distinguishes a sliced fruit on a dish from the same fruit in a bag. Capsule networks \cite{sabour2017dynamic} replace scalar-in/scalar-out units with vector-in/vector-out capsules and replace pooling with dynamic routing, encoding position, scale and alignment between local fragments and the whole object. In addition, the design, application, and assessment of our system provide the following research highlights:
\begin{itemize}
    \item \textbf{Dilated Fruit Capsules.} Substituting dilated for standard convolutions in the capsule front end expands the receptive field exponentially at constant kernel size, so each capsule aggregates multi-scale context before routing. Section \ref{sec:ablation} isolates this factor from the rest of the architecture.
    \item \textbf{Bayesian hyper-parameter search} over the margin-loss constants, dilation factor and training schedule, replacing hand tuning of a loss with four interacting free parameters.
    \item \textbf{PD-19}, a 19-class, 10{,}639-image in-the-wild benchmark with multiple categories per image, inhomogeneous backgrounds and occlusion, on which the strongest transfer-learning baseline loses $2.7\%$ to FruitCapsNet; and Grad-CAM evidence, propagated from DigitCaps, that the margin is attributable to whole-object rather than edge-localised attribution.
\end{itemize}
The workflow is shown in Fig. \ref{fig_pipeline}. Section \ref{sec:related} covers prior art, Section \ref{sec:method} the model, Section \ref{sec:exp} experiments, Section \ref{sec:concl} concludes.
\section{Related Work}
\label{sec:related}
\textbf{Hand-crafted descriptors with shallow classifiers} account for most of the pre-deep literature. Colour and texture statistics, reduced by PCA, feed SVMs and feed-forward networks \cite{Ref_3},\cite{Ref_4},\cite{pujari2013reduced}; Rocha \textit{et al.} \cite{Ref_6} combine Unser texture, colour-coherence vectors and border/interior descriptors across LDA, SVM, trees and NNs; shape-only pipelines use bounding-box and convex-hull geometry \cite{Ref_13},\cite{Ref_9}. Grading variants add rule-based or fuzzy decision layers over hue and brown-area measurements for banana ripeness \cite{Ref_1} and mango quality \cite{Ref_7}, edge plus colour co-occurrence for citrus \cite{pydipati2006identification}, Gabor decomposition \cite{zhang2014fruit} and active contours with fuzzy $k$-means \cite{ghabousian2012segmentation} for apples, sum-and-difference histograms for texture \cite{Ref_8}, and $k$-means over shallow features for infected-patch detection \cite{Ref_16}. Ensembles of global colour histograms, CCV, LBP and Zernike moments support robotic picking \cite{Ref_17}, and handwriting descriptors have been transferred to fruit \cite{Ref_18}. These methods are competitive only where the background is controlled.

\textbf{Non-visible sensing} trades cost for invariance: hyperspectral imaging with PCA for kiwi bruising \cite{Ref_11}, visible spectroscopy for longan \cite{Ref_12}, hyperspectral reflectance/transmittance with feature selection for cucumber chilling injury \cite{Ref_14}, geometric features for cucumber quality \cite{Ref_15}, 19-band multispectral imaging for strawberry ripeness \cite{Ref_19}, terahertz time-domain spectroscopy \cite{ren2019machine}, colour-space comparisons for Cape gooseberry \cite{castro2019classification}, and genetic-algorithm defect detection in apples \cite{zhang2020detection}. None is deployable at retail cost.

\textbf{Deep models.} Hossain \textit{et al.} \cite{hossain2018automatic} give the reference CNN and fine-tuned VGG-16 \cite{simonyan2014very} results on SMP and FruitsGB and remain the standard comparison. Capsule networks have since been applied to high-resolution scene classification \cite{guo2020deep}, remote-sensing scenes with multi-convolutional capsules \cite{raza2020diverse}, bogie fault diagnosis \cite{Explanation_1} and apoptosis classification \cite{mobiny2019automated}, but always with standard convolutions in the capsule front end. Dilated convolutions \cite{yu2016multiscale} have not previously been placed inside capsule layers, which is the gap this work addresses.
\begin{figure}[!t]
\centering
\includegraphics[width=4.9cm, height=6.6cm]{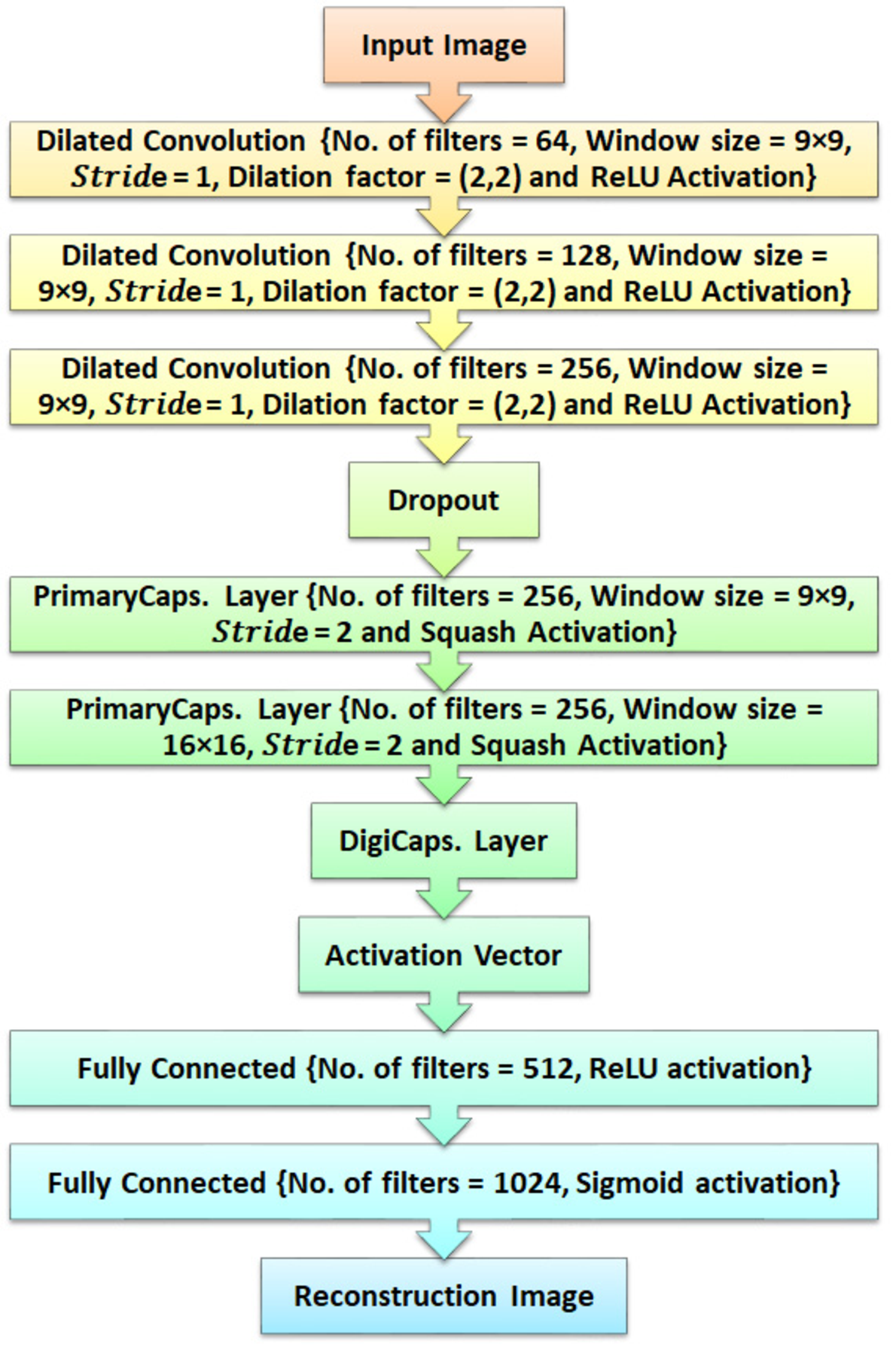}
\caption{Modular schematic of FruitCapsNet. Three dilated convolutional layers feed PrimaryCaps; DigitCaps emits one capsule per class; the decoder reconstructs the input as a regulariser.}
\label{Capsule_Network_Architecture}
\vspace{-3mm}
\end{figure}
\section{Method}
\label{sec:method}
\subsection{Architecture}
FruitCapsNet is an encoder-decoder (Fig. \ref{Capsule_Network_Architecture}). The encoder stacks three \emph{dilated} convolutional layers, a PrimaryCaps layer and a DigitCaps layer whose capsule count equals the class count, mapped through three fully connected layers. Feature-vector magnitudes are normalised by the squash non-linearity rather than element-wise scaling, and dynamic routing carries information between capsule layers. The decoder reconstructs the input from the encoder output and acts purely as a regulariser against over-fitting.

\subsection{Dilated Fruit Capsules}
A capsule is a group of neurons whose activity vector encodes the instantiation parameters of one class: its length is the class likelihood and its direction carries pose, deformation and albedo \cite{sabour2017dynamic}. This is the substantive difference from a CNN feature map, which retains neither pose nor part-whole structure \cite{guo2020deep}. Each capsule holds a weight matrix $W_{ij}$ and a routing coefficient $c_{ij}$; the $j^{th}$ parent capsule in layer $l{+}1$ receives $\tilde{u}_{j|i} = W_{ij}u_i$ from every capsule $i$ in layer $l$, with $W_{ij}$ encoding the part-whole spatial relationship \cite{raza2020diverse}.

\begin{figure}[!t]
\centering
\includegraphics[width=\columnwidth, height=2.6cm]{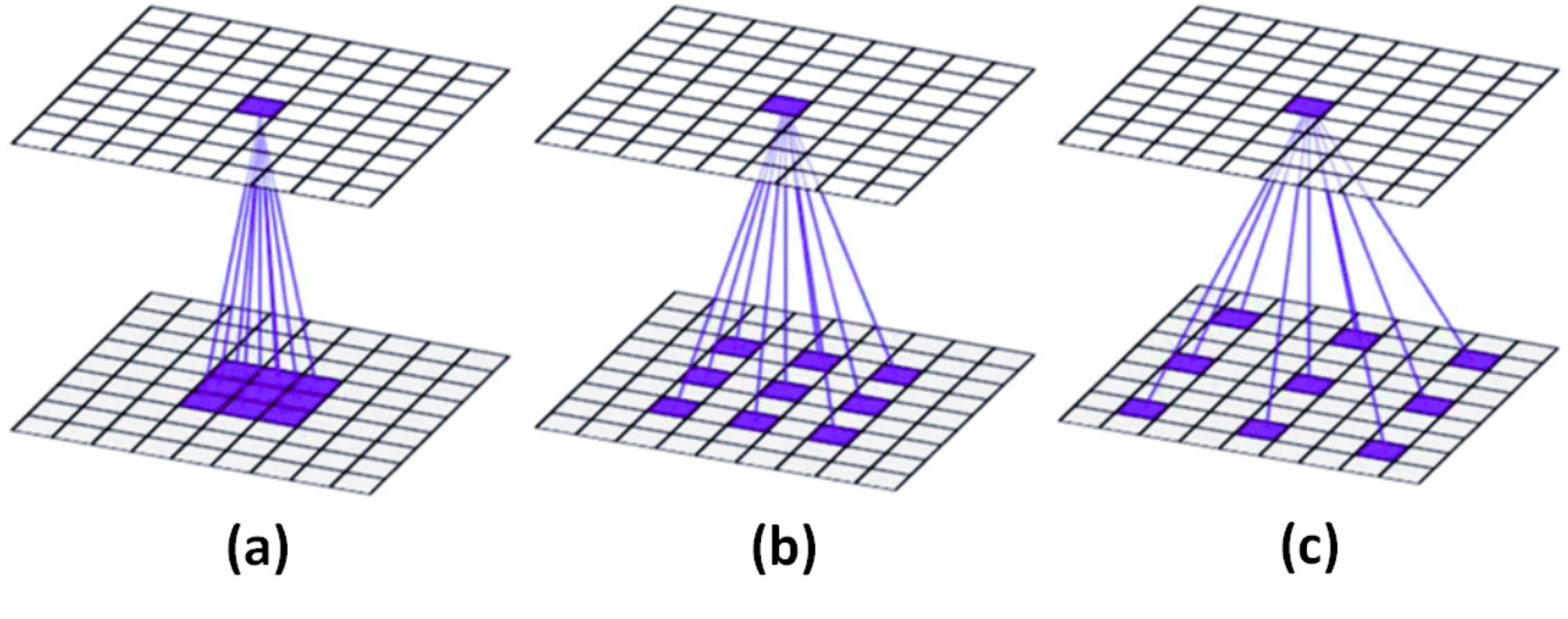}
\caption{Receptive field for dilation (a) $d=1$, (b) $d=2$, (c) $d=3$.}
\label{fig_DC_2}
\vspace{-3mm}
\end{figure}

The front end of \cite{sabour2017dynamic} uses standard convolutions, so a capsule sees only a $(2r{+}1)^2$ neighbourhood and multi-scale context must be bought with depth or stride, the latter at the cost of resolution. We substitute dilated convolutions. For kernel $k$, a standard convolution gives $(C*k)(p)=\sum_{i+j=p}C(i)k(j)$, whereas a dilated convolution with factor $d$ skips pixels, $(C_d*k)(p)=\sum_{i+dj=p}C(i)k(j)$, expanding the receptive field exponentially in $d$ at fixed parameter count and fixed resolution \cite{yu2016multiscale} (Fig. \ref{fig_DC_2}). Each Fruit Capsule therefore aggregates context across scales \emph{before} routing decides part-whole agreement, so the routing operates on evidence that already spans the fruit and its immediate background rather than a local patch. This is what we expect to matter under the occlusion and multi-object conditions of PD-19, and Section \ref{sec:ablation} tests it directly.
\subsection{Routing and Objective}
Because capsule outputs are vectors, pooling is replaced by dynamic routing \cite{sabour2017dynamic}. The input to parent capsule $j$ is $s_j=\sum_i c_{ij}\tilde{u}_{j|i}$ with $c_{ij}=\mathrm{softmax}_k(b_{ij})$, and the logits are refined over iterations as $\tilde{b}_{ij}=b_{ij}+v_j\!\cdot\!\tilde{u}_{j|i}$, so agreement between a child prediction and the parent output amplifies that child's contribution and suppresses its influence on competing parents \cite{Explanation_1}. Activations use squashing, $v_j=\|s_j\|_2 s_j / \big((1+\|s_j\|_2)\|s_j\|\big)$, which ties vector length to presence probability while preserving orientation.

Training minimises margin loss $L_j$ from the encoder plus reconstruction loss $L_r$ from the decoder,
\begin{equation}
    L_{Total} = L_j+\alpha L_r,
\label{Total_loss_equation}
\end{equation}
\begin{equation}L^j_{margin}=
     \sum_k\begin{bmatrix}
T_j\,max\left (0,m^{+}-\|v_j\|  \right )^2\\
+\lambda\left (1-T_j  \right )max\left (0,\|v_j\|-m^{-} \right )^2
\end{bmatrix}
\label{loss_equation}
\end{equation}
where $T_j\!=\!1$ if class $j$ is present and $\alpha$ prevents the decoder from dominating.

\subsection{Bayesian Hyper-parameter Search}
Eqs. \ref{Total_loss_equation}-\ref{loss_equation} introduce four interacting constants ($\alpha, m^{+}, m^{-}, \lambda$) on top of the usual schedule, and the validation error is non-differentiable and expensive to evaluate \cite{Bayesian_2}. We therefore use Bayesian optimisation with a Gaussian-process surrogate \cite{Bayesian_1},\cite{Bayesian_3} over those four constants plus depth, filter size, kernel parameters, \emph{dilation factor}, learning rate, batch size, dropout, $L_2$ strength and the ADAM moment decay rates, with integer or log-scale search intervals declared per variable. The selected values are $\alpha\!=\!5\!\times\!10^{-4}$, $m^{+}\!=\!0.4$, $m^{-}\!=\!0.2$, $\lambda\!=\!0.7$, ADAM \cite{kingma2014adam} at $10^{-3}$, batch 32, dropout $5\%$, $L_2\!=\!0.1$ \cite{srivastava2014dropout},\cite{van2017l2}; channel and filter configurations are in Fig. \ref{Capsule_Network_Architecture}.
\section{Experiments}
\label{sec:exp}
\subsection{Datasets}
Table \ref{tab:datasets} summarises the four benchmarks; per-class counts are in the supplementary material. \textbf{SMP} \cite{rocha2010automatic} was collected at a distribution centre under varying illumination and pose, with specular reflection, shadow and occlusion from plastic-bag packaging. \textbf{FruitsGB} \cite{FRUITSGB_dataset} pairs six Indian fruit varieties with a good/bad quality label, giving 12 classes photographed on a phone camera against diverse backgrounds. \textbf{Fruits-360} \cite{Fruits_360_dataset},\cite{murecsan2018fruit} is pre-segmented: single fruits on a rotating shaft against a white background, free of illumination variation.

\begin{figure}[!t]
\centering
\includegraphics[width=\columnwidth, height=3.9cm]{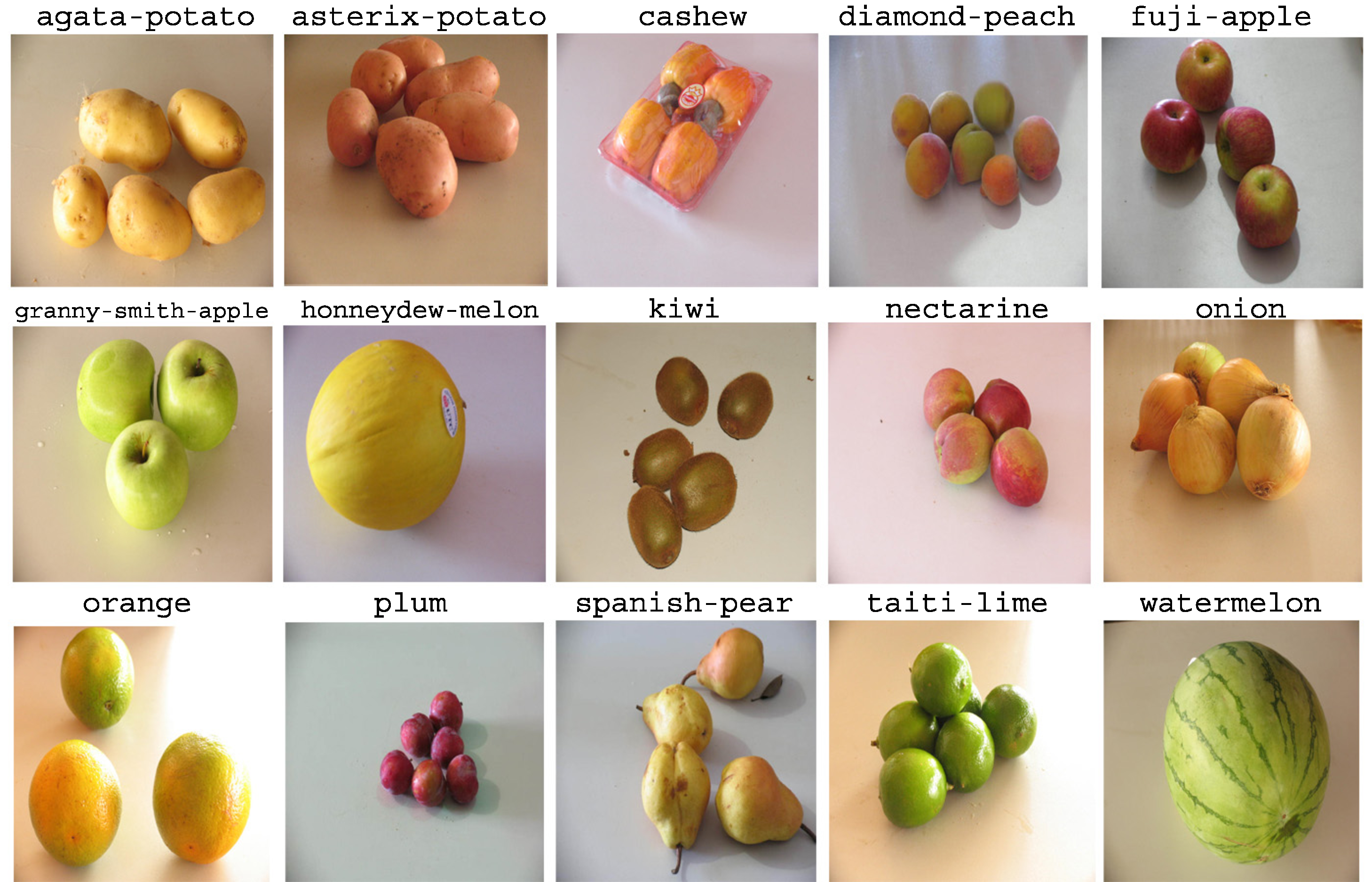}
\caption{SMP samples: illumination, pose and count vary within class.}
\label{Supermarket_figure}
\vspace{-3mm}
\end{figure}

\begin{figure}[!t]
\centering
\includegraphics[width=\columnwidth, height=3.9cm]{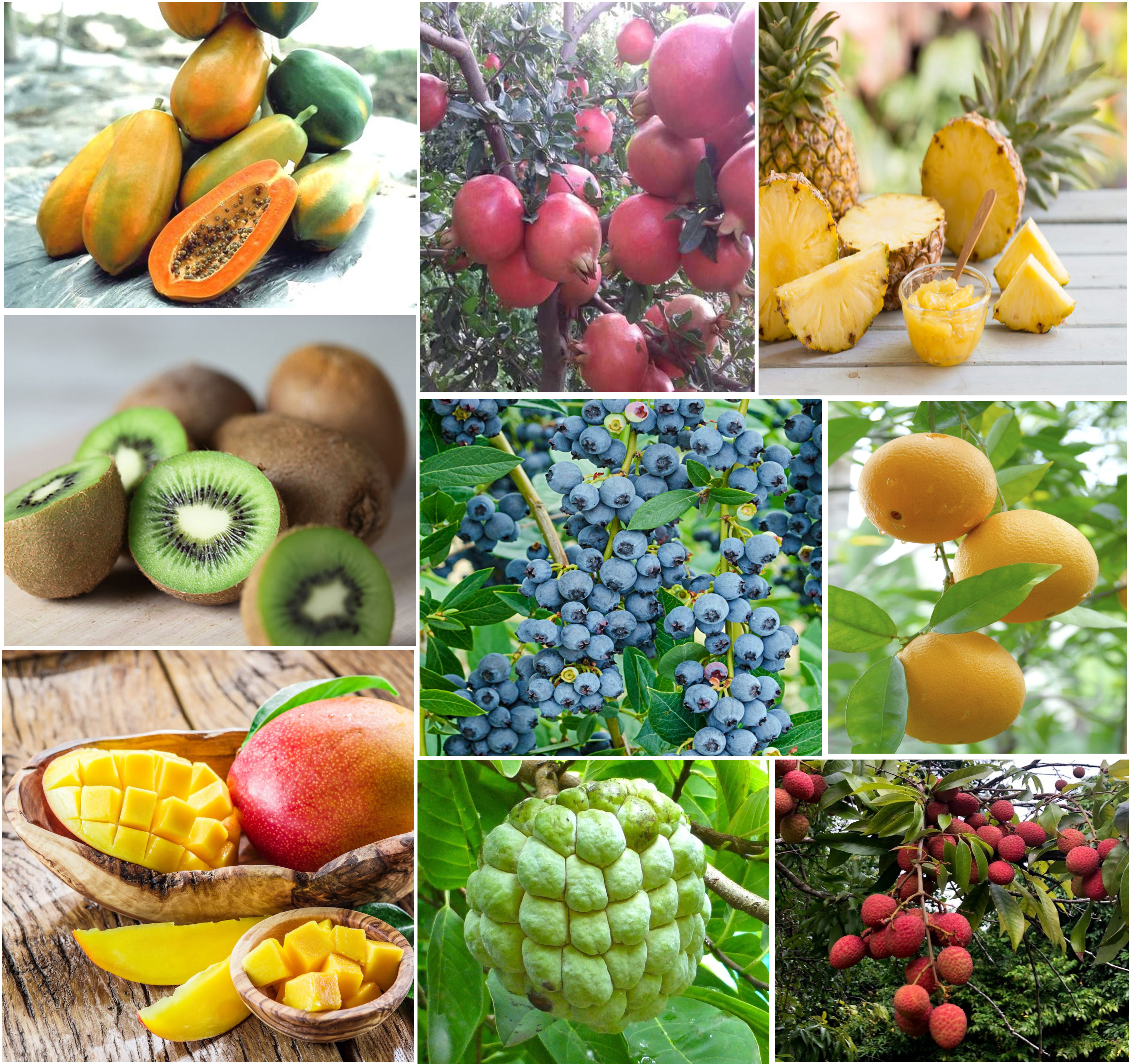}
\caption{PD-19 samples: multiple categories per image, cluttered backgrounds, occlusion, and cut/plated/unpicked presentations.}
\label{PD_19_images}
\vspace{-3mm}
\end{figure}

All three contain one fruit class per image on a broadly homogeneous background and are close to saturated. \textbf{PD-19} was assembled to break that assumption: 10{,}639 web images over 19 classes containing multiple categories per image, inhomogeneous backgrounds, diverse illumination, and fruit uncut, halved, bagged, plated or unpicked (Figs. \ref{Supermarket_figure}, \ref{PD_19_images}). PD-19 is the discriminative benchmark in what follows.
\begin{table}[!t]
\centering
\caption{Datasets. Per-class breakdowns in supplementary material.}
\label{tab:datasets}
\scalebox{0.86}{
\begin{tabular}{@{}|l|c|c|c|l|@{}}
\midrule
\textbf{Dataset}&\textbf{Cls.}&\textbf{Images}&\textbf{Res.}&\textbf{Character}\\
\midrule
SMP \cite{rocha2010automatic}&15&2{,}633&$1024{\times}768$&bagged, shadowed, multi-count\\
\midrule
FruitsGB \cite{FRUITSGB_dataset}&12&12{,}000&$256{\times}256$&6 fruits $\times$ good/bad quality\\
\midrule
Fruits-360 \cite{murecsan2018fruit}&81&55{,}244&$100{\times}100$&pre-segmented, white bg.\\
\midrule
\textbf{PD-19 (ours)}&19&10{,}639&variable&\textbf{in the wild, multi-class/image}\\
\midrule
\end{tabular}}
\vspace{-2mm}
\end{table}
\subsection{Protocol}
FruitCapsNet is implemented in TensorFlow/Keras on Ubuntu 20.04, AMD Ryzen 7 3700X, 32\,GB DDR4, NVIDIA RTX 3080 (10\,GB). SMP, FruitsGB and PD-19 are split $0.6\!:\!0.2\!:\!0.2$ for train/validation/test; splits are redrawn several times and mean test metrics reported. 
For Fruits-360 we use the official 41{,}322/13{,}877 train/test partition, since consecutive frames of the same rotating specimen make random splitting leakage-prone. Images are resized per dataset subject to memory limits. Performance is reported as macro-averaged precision, recall and $F_1$ plus Cohen's $\kappa$, defined as in \cite{Explanation_1}.

\begin{figure*}[!htp]
\centering
\includegraphics[width=\textwidth, height=3.9cm]{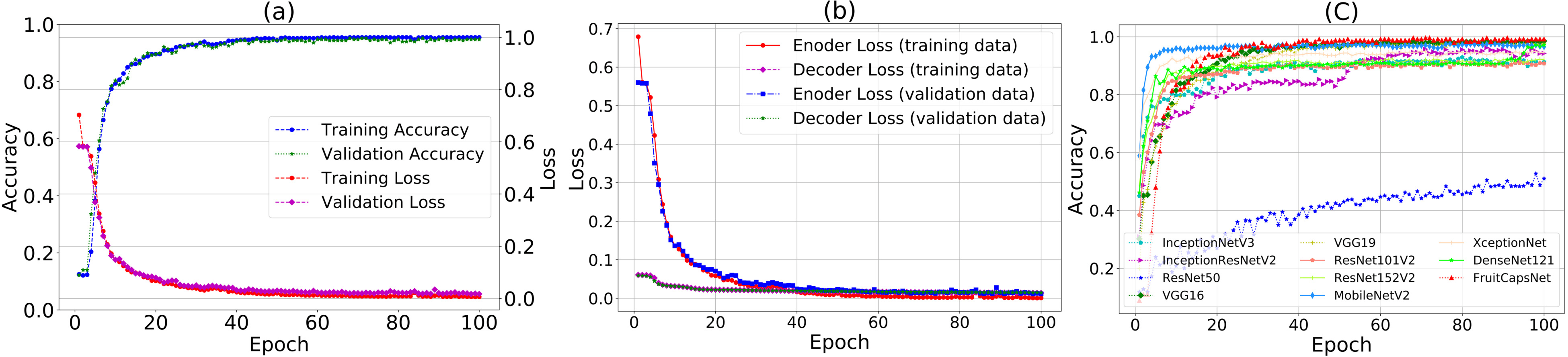}
\caption{Over the epochs, variation in FruitCapsNet‘s (a) accuracy and total loss, (b) different model loss components (encoder \& decoder loss), and (c) comparison of network convergence for different TL architectures for the SMP dataset.}
\label{fig_SMP_14_class_Training_Epoch}
\vspace{-2mm}
\end{figure*}

\begin{figure*}[!htp]
\centering
\includegraphics[width=\textwidth, height=4.2cm]{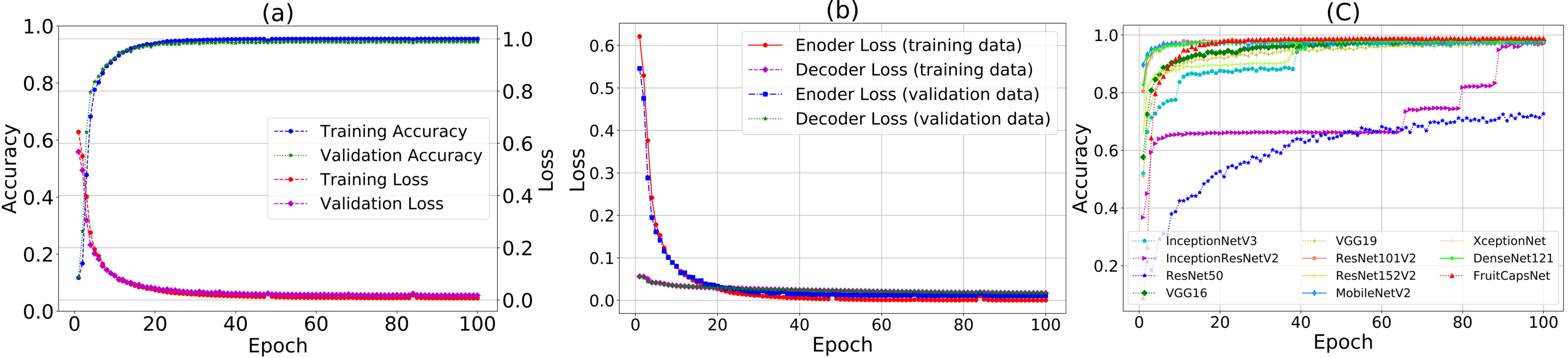}
\caption{Over the epochs, variation in FruitCapsNet‘s (a) accuracy and total loss, (b) different model loss components (encoder \& decoder loss), and (c) comparison of network convergence for different TL architectures for the FruitsGB dataset.}
\label{fig_FruitsGB_12_class_Training_Epoch}
\vspace{-2mm}
\end{figure*}

\begin{figure*}[!htp]
\centering
\includegraphics[width=\textwidth, height=4.2cm]{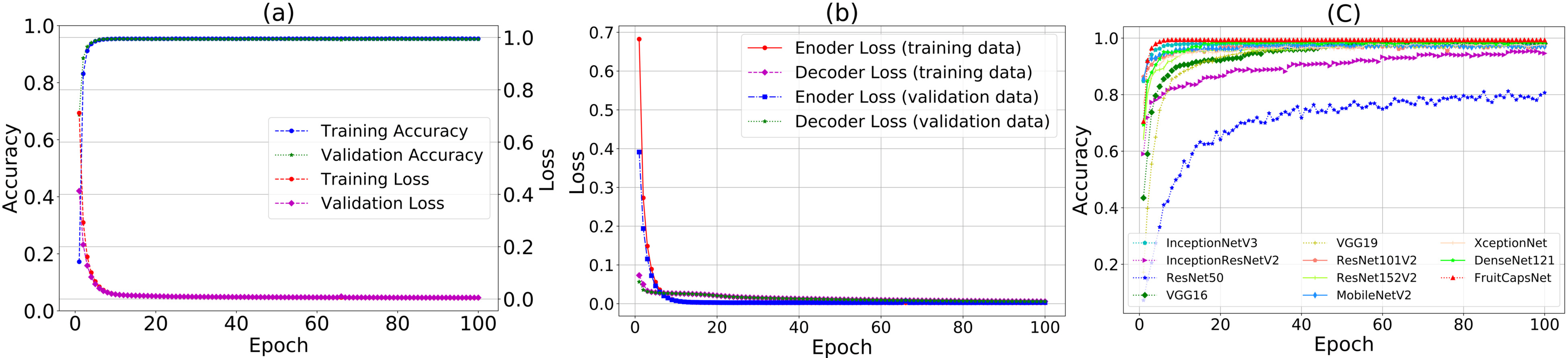}
\caption{Over the epochs, variation in FruitCapsNet‘s (a) accuracy and total loss, (b) different model loss components (encoder \& decoder loss), and (c) comparison of network convergence for different TL architectures for the Fruit 360 dataset.}
\label{fig_Fruits360_81_class_Training_Epoch}
\vspace{-2mm}
\end{figure*}

\begin{figure*}[!htp]
\centering
\includegraphics[width=\textwidth, height=4.2cm]{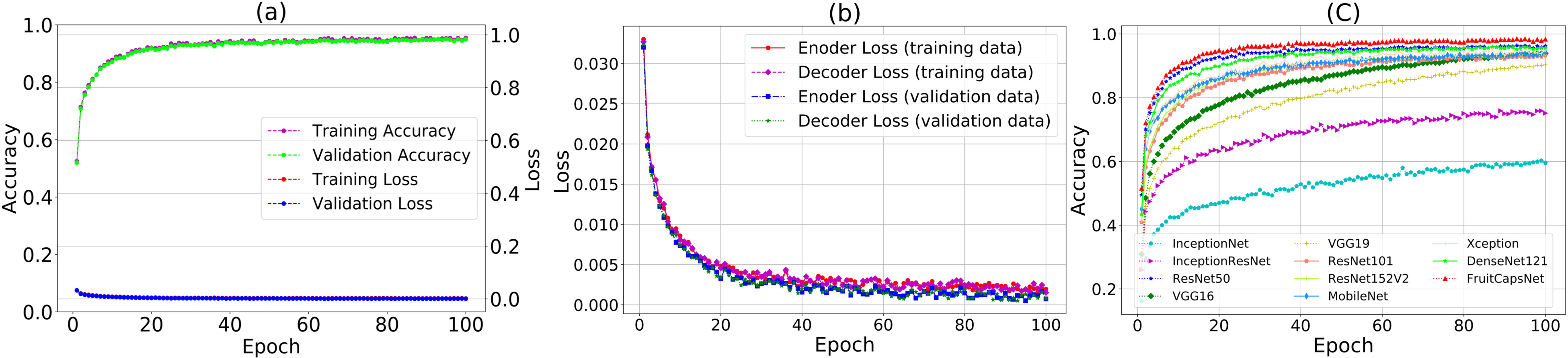}
\caption{Over the epochs, variation in FruitCapsNet‘s (a) accuracy and total loss, (b) different model loss components (encoder \& decoder loss), and (c) comparison of network convergence for different TL architectures for the PD-19 dataset.}
\label{fig_PD19_19_class_Training_Epoch}
\vspace{-2mm}
\end{figure*}

The accuracy, total loss, and the encoder/decoder loss components over 100 epochs on SMP (Fig. \ref{fig_SMP_14_class_Training_Epoch}), FruitsGB (Fig. \ref{fig_FruitsGB_12_class_Training_Epoch}), Fruits-360 (Fig. \ref{fig_Fruits360_81_class_Training_Epoch}), and PD-19 (Fig. \ref{fig_PD19_19_class_Training_Epoch}), together with convergence of the transfer-learning baselines is also illustrated. Behaviour on SMP, FruitsGB and Fruits-360 is qualitatively identical and is omitted for space; steady-state reconstruction loss rises with class count, as expected \cite{Explanation_1}.
\begin{figure*}[!htp]
\centering
\includegraphics[width=\textwidth, height=4.3cm]{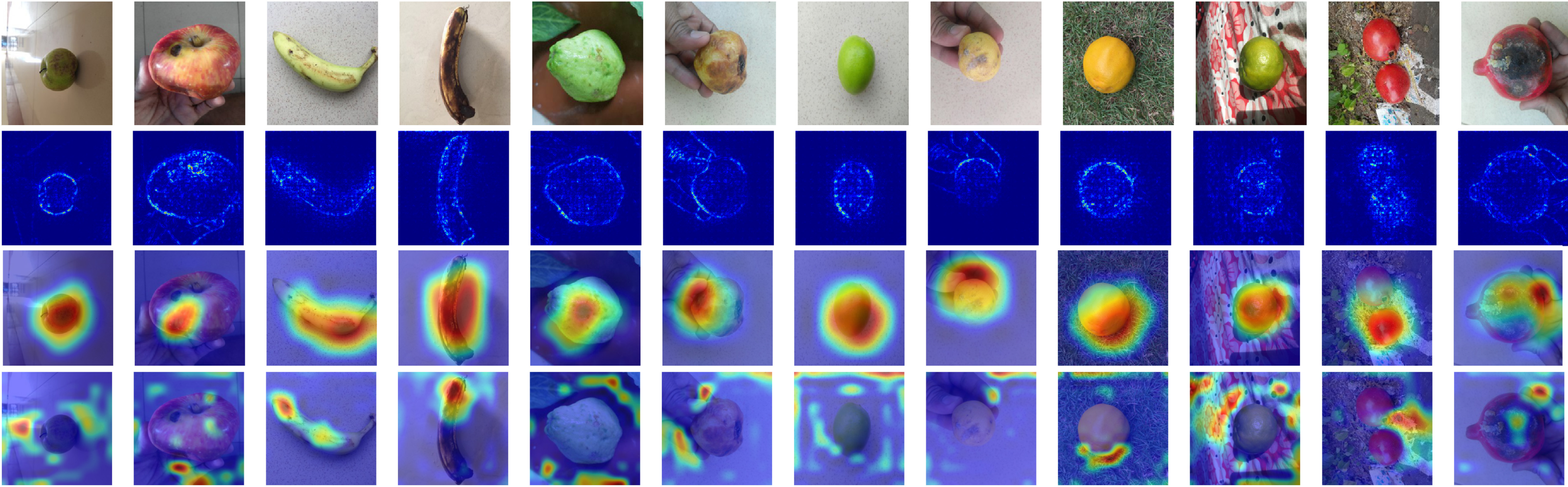}
\caption{FruitsGB: inputs ($1^{st}$ row), dense-layer maps ($2^{nd}$), Grad-CAM for FruitCapsNet ($3^{rd}$) and DenseNet ($4^{th}$)..}
\label{GradCam_FruitsGB}
\end{figure*}
\begin{figure*}[!t]
\centering
\includegraphics[width=\textwidth, height=4.3cm]{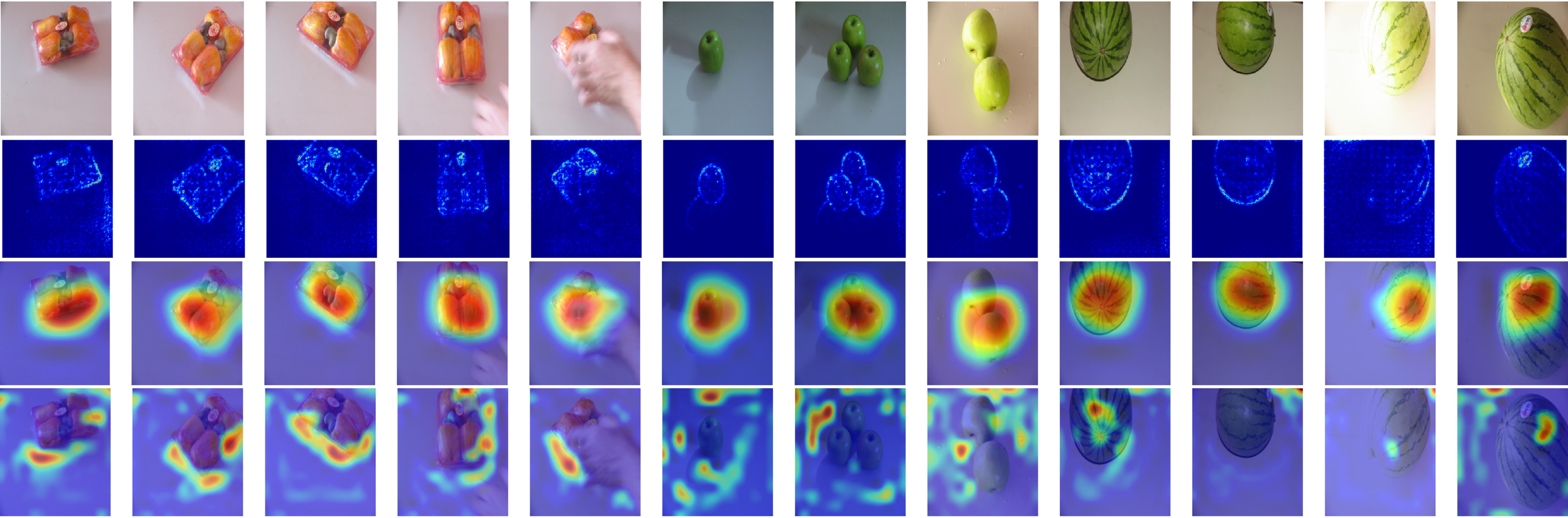}
\caption{SMP: inputs ($1^{st}$ row), dense-layer maps ($2^{nd}$), Grad-CAM for FruitCapsNet ($3^{rd}$) and DenseNet ($4^{th}$).}
\label{GradCam_Supermarket}
\vspace{-2mm}
\end{figure*}
\begin{figure*}[!t]
\centering
\includegraphics[width=\textwidth, height=4.3cm]{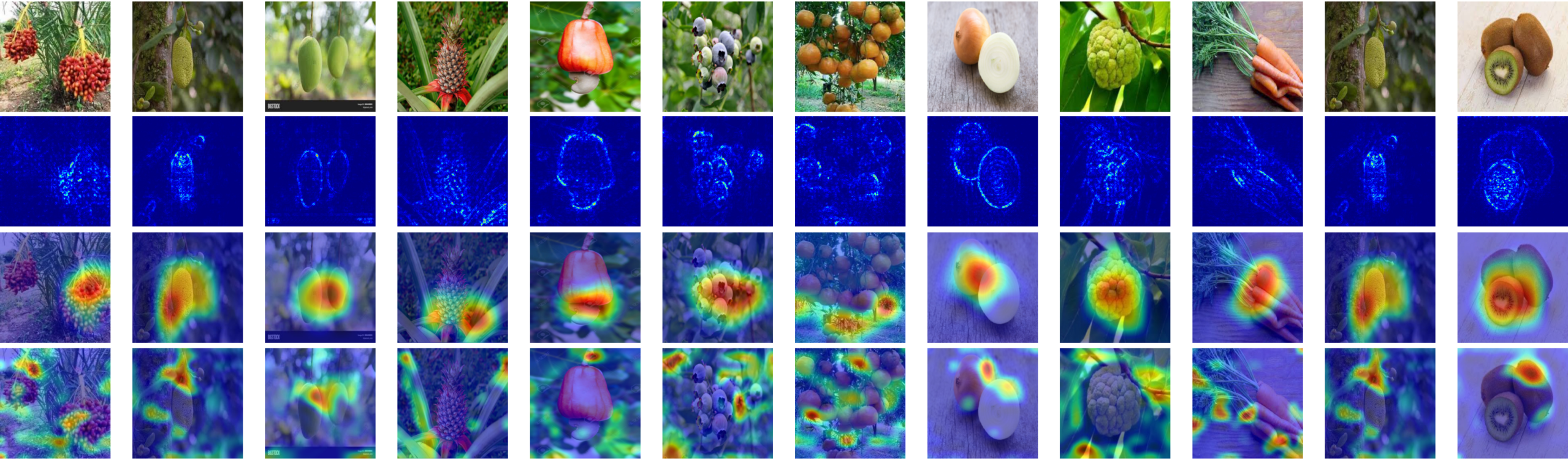}
\caption{PD-19: inputs ($1^{st}$ row), dense-layer maps ($2^{nd}$), Grad-CAM for FruitCapsNet ($3^{rd}$) and DenseNet ($4^{th}$). FruitCapsNet retains whole-object attribution under occlusion and clutter.}
\label{GradCam_PD_19}
\vspace{-2mm}
\end{figure*}

\subsection{Results}
Table \ref{Fruit_classification_results} gives mean test metrics over repeated splits. $\kappa>0.98$ on all four datasets indicates near-perfect agreement with ground truth, including on PD-19 where the class prior is uneven.

Table \ref{Pre_trained_Models} compares against ten fine-tuned ImageNet backbones \cite{ResNet},\cite{VGG},\cite{DenseNet},\cite{InceptionNet},\cite{Inception_ResNet},\cite{Xception},\cite{MobileNet},\cite{russakovsky2015imagenet}, with deeper layers unfrozen for the AFCW task. DenseNet is the nearest competitor. FruitCapsNet is best on every dataset at depth 7 versus 201, with comparable parameter count, and the margin scales with difficulty: $+0.80\%$, $+0.32\%$ and $+0.90\%$ on SMP, FruitsGB and Fruits-360, but $+2.67\%$ on PD-19. That ordering is the substantive result; the three public sets are saturated and differences there are within run-to-run noise. Table \ref{SOTA_1} places FruitCapsNet against published results per dataset.
\begin{table}[!t]
\centering
\caption{FruitCapsNet test performance (\%), mean over repeated splits.}
\label{Fruit_classification_results}
\scalebox{0.95}{
\begin{tabular}{@{}|l|c|c|c|c|c|@{}}
\midrule
\textbf{Dataset}&\textbf{Acc.}&$\pmb{F_1}$&\textbf{Rec.}&\textbf{Prec.}&$\pmb{\kappa}$\\
\midrule
FruitsGB&98.84&98.54&98.98&98.17&98.72\\
\midrule
SMP&99.19&99.39&99.02&99.49&99.12\\
\midrule
Fruits-360&99.31&99.28&99.41&99.07&99.28\\
\midrule
PD-19&98.28&98.25&98.55&97.91&98.43\\
\midrule
\end{tabular}}
\vspace{-2mm}
\end{table}
\begin{table}[!t]
\centering
\caption{Comparison with fine-tuned pre-trained backbones (accuracy, \%).}
\label{Pre_trained_Models}
\scalebox{0.82}{
\begin{tabular}{@{}|l|c|c|c|c|c|c|@{}}
\midrule
\textbf{Network}&\textbf{Depth}&\textbf{PM}&\textbf{SMP}&\textbf{FruitsGB}&\textbf{Fruits-360}&\textbf{PD-19}\\
\midrule
ResNet-50 \cite{ResNet}&50&25.63&49.86&72.02&80.28&89.01\\
\midrule
ResNet-101 \cite{ResNet}&101&44.70&90.85&97.16&97.20&93.05\\
\midrule
ResNet-152 \cite{ResNet}&152&60.38&92.18&98.05&96.96&93.51\\
\midrule
VGG-16 \cite{VGG}&16&138.3&96.91&97.77&98.00&93.84\\
\midrule
VGG-19 \cite{VGG}&19&143.6&97.46&96.97&97.58&91.22\\
\midrule
DenseNet-201 \cite{DenseNet}&201&20.0&98.39&98.52&98.41&95.61\\
\midrule
Inception \cite{InceptionNet}&48&23.85&91.12&97.36&97.91&89.88\\
\midrule
Inception-ResNet \cite{Inception_ResNet}&164&55.87&94.25&96.91&94.70&94.50\\
\midrule
Xception \cite{Xception}&71&22.91&94.52&98.16&97.60&92.65\\
\midrule
MobileNet \cite{MobileNet}&53&3.53&96.59&97.66&97.00&90.95\\
\midrule
\textbf{FruitCapsNet}&\textbf{7}&21.61&\textbf{99.19}&\textbf{98.84}&\textbf{99.31}&\textbf{98.28}\\
\midrule
\end{tabular}}
\begin{tablenotes}\footnotesize
\item[*] PM = trainable parameters (millions).
\end{tablenotes}
\vspace{-2mm}
\end{table}
\begin{table}[!t]
\centering
\caption{Comparison with published results. Prior art on SMP and Fruits-360 spans hand-crafted descriptors and CNN detectors; FruitsGB and PD-19 have a single deep baseline.}
\label{SOTA_1}
\scalebox{0.82}{
\begin{tabular}{@{}|l|l|l|c|@{}}
\midrule
\textbf{Set}&\textbf{Ref}&\textbf{Method}&\textbf{Acc.}\\
\midrule
\multirow{5}{*}{SMP}
&\cite{comparison_1}&colour/shape/texture + wavelet co-occurrence&86.00\\
\cmidrule{2-4}
&\cite{comparison_2}&autocorrelogram, CCV, BIC, LAS + ML fusion&98.80\\
\cmidrule{2-4}
&\cite{comparison_3}&LBP, HOG, GaborLBP, CNN+SVM, R-CNN&98.50\\
\cmidrule{2-4}
&\cite{hossain2018automatic}&CNN, VGG-16&88.35\\
\cmidrule{2-4}
&\textbf{ours}&\textbf{FruitCapsNet + BO}&\textbf{99.19}\\
\midrule
\multirow{6}{*}{\begin{tabular}[c]{@{}l@{}}Fruits\\-360\end{tabular}}
&\cite{comparison_7}&pure CNN, global average pooling&98.88\\
\cmidrule{2-4}
&\cite{comparison_5}&YOLO / YOLOv3 / improved Faster R-CNN&70.14-90.73\\
\cmidrule{2-4}
&\cite{comparison_8}&CNN + EfficientNet&95.67\\
\cmidrule{2-4}
&\cite{comparison_9}&MobileNet-V2, ShuffleNet-V2, LeNet&98.70\\
\cmidrule{2-4}
&\cite{comparison_6}&LBP, Hu moments, HSV + random forest&94.83\\
\cmidrule{2-4}
&\cite{comparison_10}&attention DenseNet + conv. autoencoder&95.86\\
\cmidrule{2-4}
&\textbf{ours}&\textbf{FruitCapsNet + BO}&\textbf{99.31}\\
\midrule
\multirow{2}{*}{FruitsGB}&\cite{hossain2018automatic}&CNN, VGG-16&90.57\\
\cmidrule{2-4}
&\textbf{ours}&\textbf{FruitCapsNet + BO}&\textbf{98.84}\\
\midrule
\multirow{2}{*}{PD-19}&\cite{hossain2018automatic}&CNN, VGG-16&65.92\\
\cmidrule{2-4}
&\textbf{ours}&\textbf{FruitCapsNet + BO}&\textbf{98.28}\\
\midrule
\end{tabular}}
\vspace{-2mm}
\end{table}
\subsection{Ablation: Isolating the Dilated Front End}
\label{sec:ablation}
To evaluate FruitCapsNet's core contributions the dilated capsule front end and Bayesian hyperparameter search we isolate each component against a standard CapsNet baseline \cite{sabour2017dynamic} ($d=1$) while keeping network depth, capsule dimensions, routing iterations, and training budget constant (Table~\ref{tab:ablation}). \textbf{Effect of Dilation ($d=2$).}
Setting $d=2$ yields the largest individual gains, accounting for 50\%–62\% of total improvement on public datasets and 70\% (+1.09\%) on PD-19. The effect scales with scene difficulty: expanding the receptive field prior to routing benefits multi-object, cluttered, and occluded scenes (PD-19) significantly more than pre-segmented, isolated specimens (Fruits-360). \textbf{Non-Monotonicity ($d=3$).}
Increasing dilation to $d=3$ degrades performance by 0.19\%–0.46\% points across all datasets. Because parameter counts are fixed, this drop is non-capacity-related. Instead, gridding artifacts \cite{yu2016multiscale} disrupt fine boundary and texture details, while an overextended receptive field integrates background noise that routing must actively suppress. An intermediate factor ($d=2$) is optimal and automatically identified by the Bayesian search.

Bayesian optimization contributes an additive, dataset-agnostic boost (+0.29\% to +0.69\%), indicating that loss-constant tuning optimizes learning schedules rather than spatial context. On PD-19, where the best transfer baseline reaches 95.61\% (Table~\ref{Pre_trained_Models}), an unmodified CapsNet yields +1.11 \% of the total +2.67\% gain, while our dilated front end and Bayesian tuning provide the remaining +1.56\%.
\begin{table}[!t]
\centering
\caption{Ablation (accuracy, \%; mean over 3 seeds).}
\label{tab:ablation}
\scalebox{0.86}{
\begin{tabular}{@{}|l|c|c|c|c|@{}}
\midrule
\textbf{Configuration}&\textbf{SMP}&\textbf{FruitsGB}&\textbf{Fruits-360}&\textbf{PD-19}\\
\midrule
CapsNet \cite{sabour2017dynamic}, $d{=}1$, hand-tuned&98.32&97.46&98.54&96.72\\
\midrule
\quad + dilation $d=2$&98.81&98.15&99.02&97.81\\
\midrule
\quad + dilation $d=3$&98.56&97.92&98.83&97.35\\
\midrule
\quad + dilation $d{=}2$ + BO (full)&\textbf{99.19}&\textbf{98.84}&\textbf{99.31}&\textbf{98.28}\\
\midrule
\end{tabular}}
\vspace{-2mm}
\end{table}
\subsection{Visual Explanation}
Grad-CAM \cite{Grad_cam},\cite{XAI_2} produces class-discriminative saliency by weighting activations with the gradient of the class score, localising the regions that drive the prediction \cite{XAI_1},\cite{XAI_3},\cite{XAI_4}. We adapt it to capsules by propagating the gradient from the target DigitCaps capsule back into the convolutional capsule layer, and compare against DenseNet, the strongest baseline in Table \ref{Pre_trained_Models}. Figs. \ref{GradCam_Supermarket} and \ref{GradCam_PD_19} show inputs, PrimaryCaps dense-layer maps, and both heatmaps for SMP and PD-19; FruitsGB behaves similarly and is omitted. Two things are visible. First, FruitCapsNet attributes the decision to the whole fruit region, spanning textural interior and boundary, whereas DenseNet's saliency collapses onto object edges. Second, FruitCapsNet's attribution stays on the fruit under the conditions that define PD-19: pose change, illumination change, several fruits in frame, and cropping or occlusion, where DenseNet's heatmaps become ambiguous or drift onto background. This is the behaviour dynamic routing predicts. Pooling forwards the strongest activation and discards the rest, so a CNN's evidence is necessarily local; routing forwards whichever child capsules agree on a consistent pose for the whole object, so evidence is aggregated over the object's extent, and the dilated front end widens the context each child capsule sees before that vote is taken. The result is that accuracy degrades least exactly where spatial structure matters most, which is consistent with the $+2.67\%$ margin on PD-19 against $+0.32\%$ on FruitsGB.
\section{Conclusion}
\label{sec:concl}
Pooling-based CNNs localise evidence to small regions and discard the spatial relations between them, which limits fruit recognition in the wild. Placing dilated convolutions inside capsule layers lets each capsule aggregate multi-scale context before dynamic routing resolves part-whole agreement, and Bayesian optimisation makes the resulting six-way hyper-parameter interaction tractable. Across four benchmarks the model beats ten fine-tuned backbones at one-third the depth, and the margin grows with scene difficulty: negligible on the saturated public sets, $+2.67\%$ on PD-19. Grad-CAM through DigitCaps shows the mechanism, whole-object rather than edge-local attribution that survives occlusion, clutter and multi-object frames. Limitations remain: PD-19 is web-collected and its label noise is unquantified, and routing cost still scales with capsule count, which constrains edge deployment. Next steps are a quantitative localisation metric for the explainability claim, adversarial or self-supervised capsule pre-training, and few-shot extension to unseen cultivars.
\bibliographystyle{IEEEtran}
\bibliography{References}
\end{document}